\documentclass{article}
\usepackage[final]{nips_2018}
\usepackage[utf8]{inputenc}
\usepackage{epsfig}
\usepackage{graphicx}
\usepackage{amsmath}
\usepackage{amssymb}
\usepackage{booktabs}
\usepackage{array}
\usepackage{adjustbox}
\usepackage{caption}
\usepackage{multirow}
\usepackage{color}
\usepackage{hyperref}
\usepackage{tabularx}

\title{PatchNet: Context-Restricted Architectures \\ to Provide Visual Features for Image Classification}
\author{Adityanarayanan Radhakrishnan\\
  Laboratory for Information \& Decision Systems\\
  Institute for Data, Systems, and Society\\
  Massachusetts Institute of Technology\\  
  Cambridge, MA 02139 \\
  \texttt{aradha@mit.edu} \\
  \And 
  Charles Durham\\
  Ann Arbor, MI 48103 \\
  \texttt{cpdurham@gmail.com} \\
  \And
  Ali Soylemezoglu \\
  Massachusetts Institute of Technology\\
  Cambridge, MA, 02139 \\
  \texttt{alican@mit.edu}
  \And
  Caroline Uhler \\
  Laboratory for Information \& Decision Systems\\
  Institute for Data, Systems, and Society\\
  Massachusetts Institute of Technology\\  
  Cambridge, MA 02139 \\
  \texttt{cuhler@mit.edu} \\
  }
\date{October 2018}

\begin{document}

\maketitle

\begin{abstract}
Understanding how a complex machine learning model makes a classification decision is essential for its acceptance in sensitive areas such as health care. 
Towards this end, we present PatchNet, a method that provides the features indicative of each class in an image using a tradeoff between restricting global image context and classification error.  We mathematically analyze this tradeoff, 
demonstrate Patchnet's ability to construct sharp visual heatmap representations of the learned features, and quantitatively compare these features with features selected by domain experts by applying PatchNet to the classification of benign/malignant skin lesions from the ISBI-ISIC 2017 melanoma classification challenge. 
\end{abstract}

\section{Introduction and Related work }
\label{intro_and_rel_work}
With the success of ResNet~\citep{ResNet} in the ImageNet challenge~\citep{ImageNet}, convolutional neural networks (CNNs) have become pervasive in computer vision.  However the complex interactions between many parameters make it difficult to interpret the features these networks use in classification.  The goal of this work is to provide a method for binary classification (with class labels $0$ and $1$) that, by limiting the image context used for classification, is forced to identify as many distinguishing features of each class as possible.  

Many recent works have attempted to visualize features learned by deep neural networks by identifying images that maximally activate specific neurons or classification labels in a deep network \citep{netdissect2017,olah2017feature, olah2018the,Saliency,GuidedBackprop,DeconvNet,ExplanatoryGraph}.  One difficulty with these approaches is quantifying interpretability of these feature visualizations in relation to the overall classification decision.  The approaches most related to the one presented here in terms of visualizing and understanding how a CNN makes a classification decision are the Class Activation Mapping (CAM) model~\citep{CAM} and Grad-CAM~\citep{GradCAM}.  
Trained on ImageNet,  CAM and Grad-CAM successfully focused on sensible regions of an image for making classification decisions. 

However, as we will demonstrate in this work, there are significant limitations in using CAM or Grad-CAM on medical datasets: (a) As described by \citep{ReLUGaussian}, there are no mathematical guarantees that the last layer of filters after max pooling is consistent with the original spatial relationships in the image.  For example, in the extreme case, downsampling to size 1 x 1 in a CNN will lead to a complete loss of spatial relations with respect to the original image. (b) CAM and Grad-CAM tend to blur the regions of interest in the visualization due to upscaling low resolution filters. (c) Most importantly, as CAM and Grad-CAM (and similarly other methods such as \citep{PatchTissueClassification}) have access to global context, the underlying CNN need only identify a single feature indicative of a given class as opposed to all features of a class. This is undesirable in particular in medical applications, where the features need to be well understood and a complete representation of the relevant features could aid a medical doctor or pathologist in making an accurate clinical diagnosis.

To overcome these limitations, we present PatchNet: this method uses a CNN to provide classification decisions on small patches of an image by determining whether a given patch contains features of either class, and then averages the decisions made on all patches across an image to make a global classification decision.  In this way, PatchNet allows for a trade-off between generalization error and feature granularity: by decreasing patch size, generalization error increases since the network is limited in the amount of context used for classification, but feature granularity increases as we can visualize the learned features for small patches of an image.  Importantly, we show that PatchNet is guaranteed to identify as many features indicative of class $1$ and class $0$ in an image as possible.  

PatchNet is motivated by mean-field approximation techniques from variational inference as well as ensemble methods.  As is done in mean-field approximations, instead of learning conditional probability distributions for predicting the label of an image given the entire image, we instead learn a simpler conditional probability distribution for predicting the label given a small patch of the original image.  Now unlike a true mean-field approximation, we do not multiply the predictions for each patch to get a global prediction, but rather treat each of these patch predictions as an ensemble of smaller classifiers and average their classifications to generate a global classification decision.  This ensures that PatchNet does not only concentrate on the most indicative feature of a class in each image, but outputs as many features indicative of each class as possible.


\section{PatchNet}
\label{model}

In this section, we provide the mathematical motivation for the proposed method and present our method.  In Sections \ref{convergence_analysis} and \ref{extracting_visualizations} in the Appendix, we prove that our method must identify as many features of a specific class as possible and explain how to extract global and filter heatmaps from PatchNet to visualize the learned features.  


\textbf{Motivation and Notation:}\;\; Suppose we are given a dataset $\mathcal{D}$ consisting of a list of images $I^{(1)}, I^{(2)}, \ldots I^{(k)}$, with each $I^{(j)} \in \{0, 1, 2 \ldots N - 1 \}^{m \times  n \times  c}$, along with a corresponding list of labels $y_{I^{(1)}}, y_{I^{(2)}}, \ldots, y_{I^{(k)}}$, with each $y_{I^{(j)}} \in \{0, 1 \}$.  Feed-forward CNNs such as VGG~\citep{VGG} or ResNet~\citep{ResNet} directly estimate $\mathbb{P}_{y | I}(y | I)$, the conditional distribution given \emph{all} pixel values. Note that since we concentrate solely on binary classification, it suffices to estimate $\mathbb{P}_{1 | I} (1 | I)$. In order to obtain a visual representation of the learned features, instead of using all pixel values, we here propose to estimate the conditional probability distributions of \emph{patches} of pixels separately and then average the estimates across all patches in an image to create a global estimate.  Formally, if an image is chunked into $l$ possibly overlapping patches $P^{(1)}, P^{(2)}, \ldots, P^{(l)}$ with $P^{(j)} \in \{0, 1, 2, \ldots N - 1\}^{m' \times n' \times c}$ with $m' \leq m$ and $n' \leq n$, then our model estimates $\mathbb{P}_{1 | I}(1 | I)$ as an average over all patches, i.e., $\mathbb{P}_{1 | I}(1 | I) = \frac{1}{l}\textstyle\sum_{j=1}^{l} \mathbb{Q}_{1 | P^{(j)}}(1 | P^{(j)})$, where $\mathbb{Q}$ is a single learned distribution applied to each patch.

There is an inherent tradeoff between patch size ($m', n'$), generalization error, and granularity of the learned features.  For instance, with $m' = m$ and $n' = n$, our method can match any CNN by simply letting the estimate for $\mathbb{Q}$ be the estimate determined by the CNN.  In this case, the generalization error achieved by the method is the same as that of the mimicked network, but the granularity of the learned features suffers due to the large scale of feature detection. By using small $m'$, $n'$, the distribution  $\mathbb{Q}$ is estimated across a smaller input space allowing detection of local features.  We show in Appendix \ref{convergence_analysis} that smaller patch sizes provide high granularity of the features used for classification at the expense of a potentially larger generalization error.  

\begin{figure*}
\begin{center}
\includegraphics[width=0.7\textwidth]{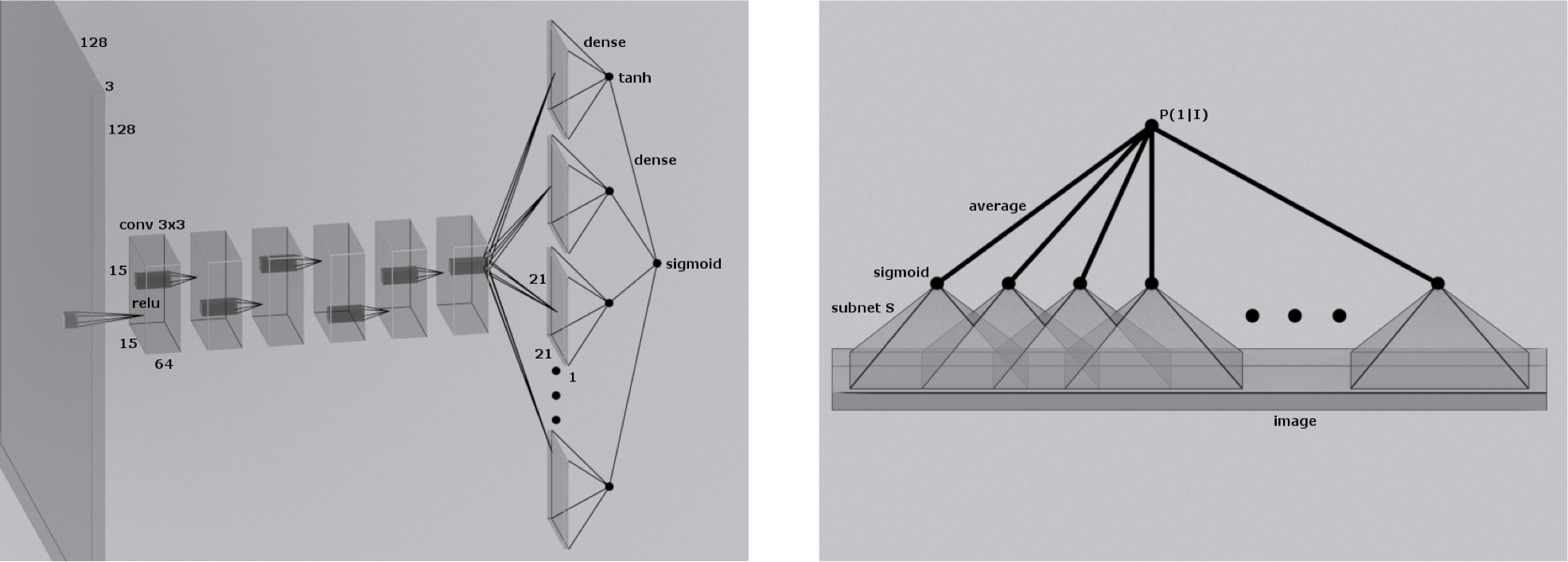}
\caption{Description of the proposed method (PatchNet). The subnet $\mathcal{S}$ that outputs a local classification decision given a patch is displayed on the left. The global component $\mathcal{G}$ consisting of repeated applications of the subnetwork $\mathcal{S}$ to provide a global classification decision given an entire image is shown on the right. The illustration uses a patch size of 15 on 128 x 128 images.}
\label{fig:Network}
\end{center}
\end{figure*}

\textbf{Method Description:}\;\; Our method consists of two components: (a) a global component $\mathcal{G}$ that outputs a global classification decision given an entire image;  (b) a local CNN $\mathcal{S}$ that outputs a local classification decision given a patch; see Figure ~\ref{fig:Network}. 

We now describe a feed-forward pass through our model. When an input image $I$ is fed into $\mathcal{G}$, the image is chunked into $l$ patches $\left[ P^{(1)}, P^{(2)}, \ldots P^{(l)} \right]$ of size $m' \times n' \times c$  with each $P^{(j)} \in \{ 0, 1, \ldots 255 \}^{m' \times n' \times c}$.  Next, these patches are aggregated and sent to $\mathcal{S}$ as a mini-batch, where they are mapped to 
\begin{equation*}
\mathcal{S}\big(\big[ P^{(1)}, P^{(2)}, \ldots, P^{(l)} \big]\big) = \big[ \mathbb{Q}_{1 | P^{(1)}}(1 | P^{(1)}), \mathbb{Q}_{1 | P^{(2)}}(1 | P^{(2)}), \ldots, \mathbb{Q}_{1 | P^{(l)}}(1 | P^{(l)}) \big],
\end{equation*}
where $\mathbb{Q}_{1 | P^{(j)}}$ is determined as follows.  A CNN consisting of $7$ layers of convolutions with $64$ filters with kernel size $3$ and $1$ pixel of padding followed by ReLU activations in each layer is applied to $P^{(j)}$ to get $64$ $m' \times n'$ filter output images.  Then $64$ linear models are applied as a dot product to each of these filter output images to reduce the $m' \times n'$ outputs to size $1$, and a tanh activation is applied to each output. Lastly, a linear layer is applied as a dot product to these $64$ outputs resulting in a single value $\mathbb{\tilde{Q}}_{1 | P^{(j)}}$ for each patch, which is converted to $\mathbb{Q}_{1 | P^{(j)}}$ by a sigmoid transformation. We use a custom CNN instead of a standard architecture to avoid using pooling layers, which are not needed when classifying on small patches.  Further, we train the CNNs from scratch to avoid potential conflicts with the scale of the learned features. 
Finally, $\mathcal{G}$ averages 
the output from $\mathcal{S}$ over all $l$ patches to obtain the global classification estimate $\mathbb{P}_{1 | I}(1 | I)$. 

Given a label $y^*$ and a global prediction $\mathbb{P}_{1 | I}(1 | I)$ we use the \emph{binary cross entropy loss} to determine the loss, namely
$L(I, y^*) = -y^* \log(\mathbb{P}_{1 | I}(1 | I)) - (1 - y^*) \log (\mathbb{P}_{0 | I}(0 | I)).$  
It is important to note that the weights for the CNN and the linear layers used to produce $\mathbb{Q}_{1 \mid P^{(j)}}$ are shared across all patches $P^{(j)}$.  Thus, the local network $\mathcal{S}$ used to produce $\mathbb{Q}_{1 | P^{(j)}}$ must be able to identify features for class $1$ and class $0$ on a local scale.  Furthermore, by averaging instead of using a linear layer to obtain the output $\mathbb{P}_{1 | I}(1 | I)$, we are inherently forcing $\mathcal{S}$ to independently identify as many features of class $1$ and class $0$ as possible without providing global context.    

\section{Experimental Results}
\label{exp_res}

We now analyze the performance of PatchNet in distinguishing between cancerous and benign skin lesions from the ISBI-ISIC 2017 melanoma detection challenge. Our training practices and hyperparameters are presented in Appendix \ref{training_practices}. In Sections \ref{DTD_experiments} and \ref{bj_mcf10a_experiments} in the Appendix, we also apply PatchNet to the Describable Textures Dataset \citep{Textures} and to a dataset of normal and fibrocystic human cell nuclei \citep{Adit}, where we show the tradeoff between patch size and generalization error.   

We used the training, validation, and test data from the ISBI-ISIC 2017 melanoma detection challenge \citep{ISBI}.  The training dataset contains 2000 images consisting of 1636 benign lesions and 374 melanoma; the validation dataset consists of 120 benign lesions and 40 melanoma; the test dataset consists of 483 benign lesions and 117 melanoma.  Since the original images vary in size, we rescaled all images down to 192 x 256.  When training PatchNet, CAM and Grad-CAM on this dataset, we followed the normalization and data augmentation procedures applied by top models from the challenge. Specifically, as in \citep{Matsunaga} we normalized the data using a gamma correction with a gamma value of 2.2 followed by applying color constancy \citep{ColorConstancy}. In addition, as in \citep{Menegola}, we augmented the data using 180 degree rotations and zoom-ins up to 120\%. Although the challenge allowed the use of external data (and all top models made use of a significant amount of external data), we chose not to use external data to provide reproducible results.  We used an equal number of training samples for each class  to ensure that the models were not learning to solely predict all images as class $0$.    

For this classification task, we trained PatchNet with a patch size of 21. The validation and test losses and accuracies for all models are provided in Table~\ref{Melanoma_Acc_Table}. Interestingly, PatchNet-21 generalizes better than CAM or Grad-CAM although it cannot make use of the global context.

\begin{table}[!t]
\footnotesize
  \centering
\begin{tabular}{|>{\centering}m{1.5cm}|>{\centering}m{1.5cm}|>{\centering}m{1.5cm}|>{\centering}m{1.5cm}|>{\centering\arraybackslash}m{1.5cm}|}
\hline
Model & Validation Loss & Validation Accuracy & Test Loss & Test Accuracy \\
\hline
PatchNet-21 & 0.499 & 77.3\% & 0.519 & 75.3\% \\
\hline
CAM & 0.538 & 76.7\% & 0.604 & 60.4\% \\
\hline
VGG-11 & 0.565 & 70.0\% & 0.538 & 74.2\% \\
\hline
\end{tabular}
\caption{Validation and test losses and accuracies for PatchNet-21, CAM, and VGG-11 on the ISBI-ISIC 2017 melanoma classification challenge.}
\label{Melanoma_Acc_Table}
\end{table}

The ISBI-ISIC melanoma 2017 detection challenge also provides the masks for four types of dermoscopic features, as determined by pathologists for these images, that are indicative of melanoma. The validation data with the feature masks consists of 150 images of which 60 do not contain any of the dermoscopic features. Table~\ref{Melanoma_Masks} provides the resulting values for average recall, average exact match and average AUROC. To avoid division by zero, we evaluated the average recall only on the 90 images that contained dermoscopic features. As illustrated by the high values for all three statistics, the features detected by PatchNet are highly correlated with the dermoscopic features provided by the pathologists. It is interesting to note that Grad-CAM achieves a high average exact match by highlighting most of the image as background at the expense of not identifying features that are indicative of malignant lesions as indicated by the low average recall.

\begin{table*}[!t]
\footnotesize
  \centering
\begin{tabular}{|>{\centering}m{1.8cm}|>{\centering}m{0.7cm}|>{\centering}m{0.7cm}|>{\centering}m{0.7cm}|>{\centering}m{0.7cm}|>{\centering}m{0.7cm}|>{\centering}m{0.7cm}|>{\centering}m{1.2cm}|>{\centering\arraybackslash}m{1.2cm}|>{\centering\arraybackslash}m{1.2cm}|}
\hline
Model & \multicolumn{3}{c|}{Benign Lesions} & \multicolumn{3}{c|}{Malignant Lesions} & Average Exact Match & Average Recall & Average AUROC \\
\hline
Original & 
\adjustimage{height=.5cm,valign=m}{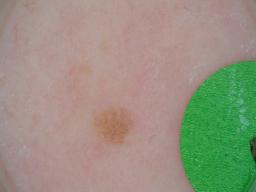} & 
\adjustimage{height=.5cm,valign=m}{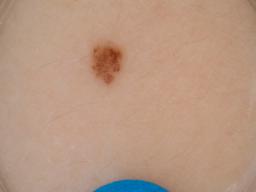} &
\adjustimage{height=.5cm,valign=m}{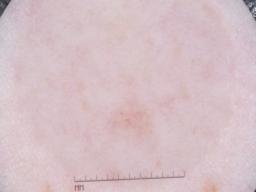} & 
\adjustimage{height=.5cm,valign=m}{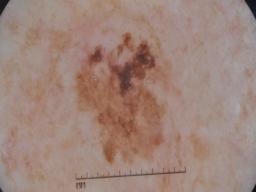} & 
\adjustimage{height=.5cm,valign=m}{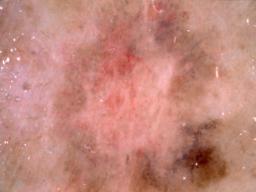} & 
\adjustimage{height=.5cm,valign=m}{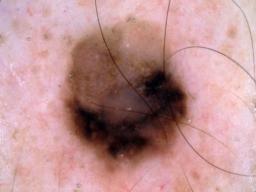} & - & - & - \\
\hline
Dermoscopic Feature Masks & 
\adjustimage{height=.5cm,valign=m}{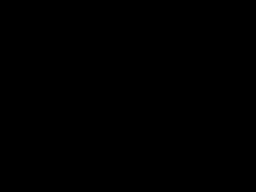} & 
\adjustimage{height=.5cm,valign=m}{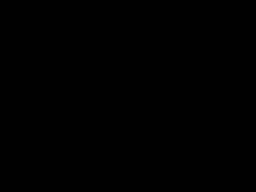} &
\adjustimage{height=.5cm,valign=m}{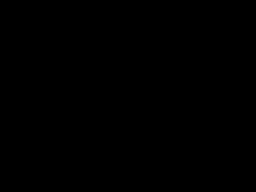} & 
\adjustimage{height=.5cm,valign=m}{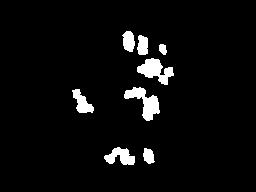} & 
\adjustimage{height=.5cm,valign=m}{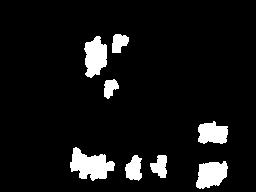} & 
\adjustimage{height=.5cm,valign=m}{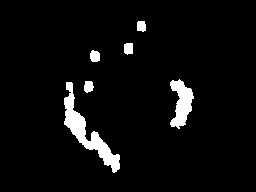} & - & - & - \\
\hline
PatchNet-21 & 
\adjustimage{height=.5cm,valign=m}{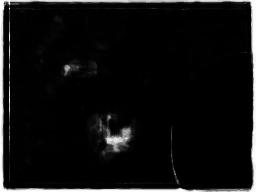} & 
\adjustimage{height=.5cm,valign=m}{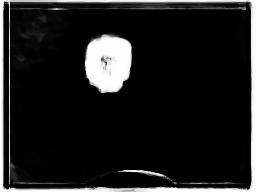} &
\adjustimage{height=.5cm,valign=m}{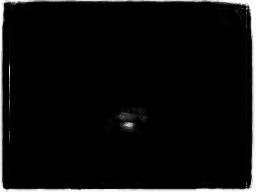} & 
\adjustimage{height=.5cm,valign=m}{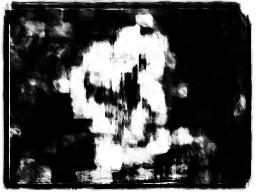} & 
\adjustimage{height=.5cm,valign=m}{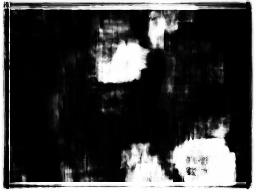} & 
\adjustimage{height=.5cm,valign=m}{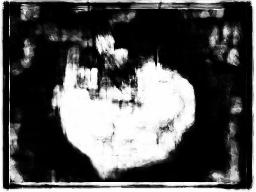} &
76.9\% & 68.8\% & 0.788 \\
\hline
CAM & 
\adjustimage{height=.5cm,valign=m}{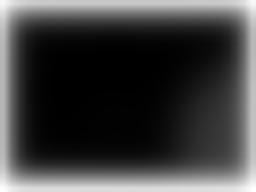} & 
\adjustimage{height=.5cm,valign=m}{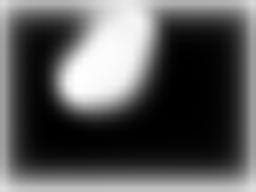} &
\adjustimage{height=.5cm,valign=m}{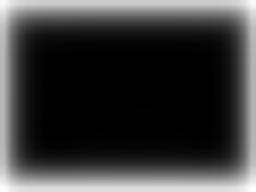} & 
\adjustimage{height=.5cm,valign=m}{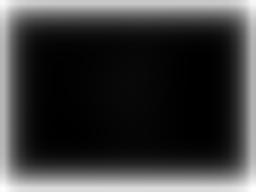} & 
\adjustimage{height=.5cm,valign=m}{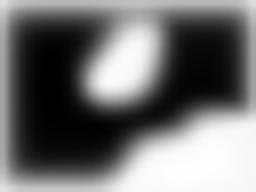} & 
\adjustimage{height=.5cm,valign=m}{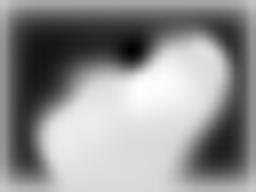} & 
62.6\% & 62.2\% & 0.667 \\
\hline
Grad-CAM & 
\adjustimage{height=.5cm,valign=m}{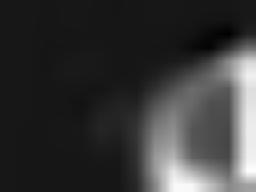} & 
\adjustimage{height=.5cm,valign=m}{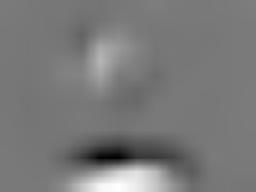} &
\adjustimage{height=.5cm,valign=m}{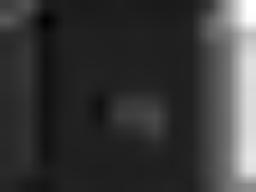} & 
\adjustimage{height=.5cm,valign=m}{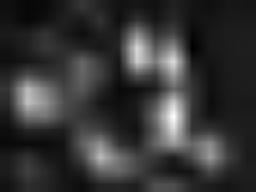} & 
\adjustimage{height=.5cm,valign=m}{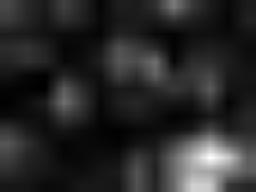} & 
\adjustimage{height=.5cm,valign=m}{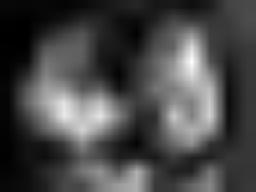} & 
87.3\% & 28.5\% & 0.725 \\
\hline
\end{tabular}
\caption{PatchNet, CAM, and Grad-CAM visualizations are compared with provided dermoscopic lesion features after the models were trained to distinguish between malignant and benign lesions.  White regions are indicative of malignant lesions.  PatchNet is able to achieve significantly higher overlap with dermascopic features provided by domain experts than CAM or Grad-CAM. }
\label{Melanoma_Masks}
\end{table*}

Table~\ref{Melanoma_Masks} also shows six sampled images from the validation set and the corresponding PatchNet-21, CAM, and Grad-CAM visualizations. As is apparent from these visualizations, PatchNet provides a more focused heatmap of the lesions than CAM or Grad-CAM. Note in particular that lesion-free skin is colored black by PatchNet, indicating that PatchNet has learned to distinguish normal skin texture as a feature of class $0$. Curiously, Grad-CAM identifies bandages (shown in green and blue on the first two images of benign lesions) as indicative of malignant lesions, thereby exemplifying the problem of using methods for medical applications that can perform classification based on only one indicative feature.

\section{Conclusions and Future Work}
\label{conclusion}
By controlling the scale of feature detection through patch size, PatchNet is able to trade-off sharpness in terms of visual heatmaps against generalization error in binary classification problems.  By overlaying the features detected by PatchNet as being indicative of melanoma with features provided by pathologists, we demonstrated the effectiveness of PatchNet to obtain medically relevant features in an unsupervised manner. We anticipate that methods such as PatchNet will in the future aid pathologists in making more accurate clinical diagnoses.   

Although PatchNet trades off generalization error in order to obtain sharp features, it achieved competitive generalization losses on all our applications.  To further improve generalization error, an interesting area for future work is to leverage both global and local context sensitive methods to achieve low generalization error while guaranteeing that the model learns from all relevant features.  
\bibliographystyle{humannat}
\bibliography{sample}

\appendix
\section{Convergence Behavior of PatchNet}
\label{convergence_analysis}
Following the notation from Section \ref{model}, the main limitation of our approach is that the patch size parameters $m', n'$ must be tuned based on the scale of features present in the dataset.  Intuitively, selecting too small values for $m', n'$ forces $\mathcal{S}$ to learn just the area of a structure as the distinguishing feature,  while selecting too large values of $m', n'$ results in complex feature interactions. For many applications, domain expertise can be used to select an appropriate patch size.

\begin{figure}[!ht]
\centering
\includegraphics[width=0.45\textwidth]{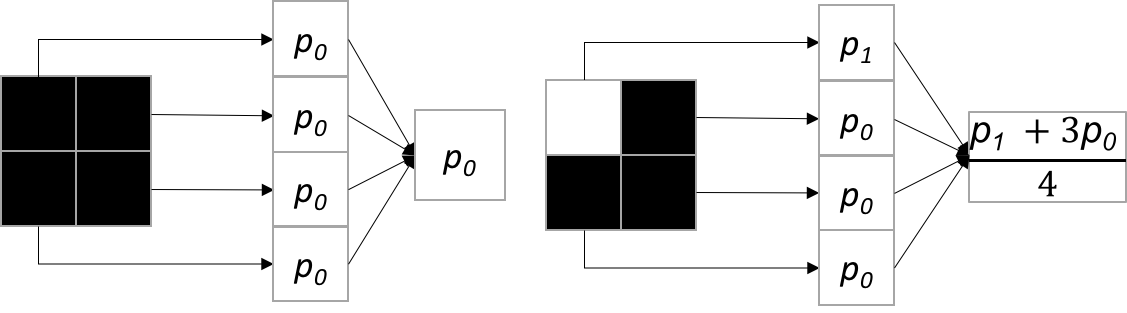}
\caption{The class $0$ images and class $1$ images are depicted on the left and right respectively.  As indicated, the images are segmented into $4$ disjoint patches.  We denote by $p_0$ the value of $\mathbb{Q}_{1 | P_0}(1 | P_0)$ for a patch $P_0$ containing all $0$-valued pixels and by $p_1$ the value of $\mathbb{Q}_{1 | P_1}(1 | P_1)$ for a patch $P_1$ containing all $1$-valued pixels.}
\label{fig:math-analysis-examples}
\end{figure}

A careful selection of the patch size is also important since it trades off feature granularity versus classification error. As illustrated by the following simple example, if an image of class 1 only contains few patches with features indicative of class $1$, then PatchNet could wrongly classify it as class $0$, since the patches $P_1$ with $\mathbb{Q}_{1 | P_1}$ close to $1$ could be out-voted by the patches $P_0$ with $\mathbb{Q}_{1 | P_0}$ close to $0$.  However, even if too few patches contain a feature of class $1$, as we show in the example below, the model is guaranteed to identify every such patch.  

\emph{Example:} Suppose that the data space $\mathcal{D}$ consists only of two images: (a) a class $0$ image that is simply an $m \times n \times 1$ image of $0$ valued pixels; (b) a class $1$ image that is an $m \times n \times 1$ image with all $0$ valued pixels except for the pixels in the upper left $\frac{m}{2} \times \frac{n}{2} \times 1$ rectangle, which all have value $1$ (see Figure~\ref{fig:math-analysis-examples}). Suppose also that our method uses $\frac{m}{2} \times \frac{n}{2} \times 1$ patches with a stride of $\frac{m}{2}$ in the first dimension and a stride of $\frac{n}{2}$ in the second dimension.  That is, the subnet is trained on $4$ patches for each class with only the upper left patch of class $1$ images containing features indicative of class $1$.

Suppose further that we train on a balanced set $\mathcal{T}=\{(I_k^{(1)}, y_{I_k^{(1)}}^*), (I_k^{(2)}, y_{I_k^{(2)}}^*), \ldots (I_k^{(T)}, y_{I_k^{(T)}}^*)\}$ of $T$ images and labels from both classes $k\in\{0, 1\}$.  Then we can deduce the values of $p_0:=\mathbb{Q}_{1 | P_0}(1 | P_0)$ and  $p_1:=\mathbb{Q}_{1 | P_1}(1 | P_1)$ for patches $P_0$ containing all $0$-valued pixels and patches $P_1$ containing all $1$-valued pixels in closed form. Given the cross entropy loss $L(I, y^*)$ for each image, the loss for all images in our training set is
\begin{equation*}
\begin{split}
L(\mathcal{T}) &=  \displaystyle\sum\limits_{i=1}^{T} L(I_0^i, y_{I_0^i}^*)  +  \displaystyle\sum\limits_{i=1}^{T} L(I_1^i, y_{I_1^i}^*) \\
&= -T\log(1 - p_0) - T\log\left(\frac{3p_0 + p_1}{4}\right). \\
\end{split}
\end{equation*}
To minimize this loss, we first analyze the derivative with respect to $p_1$, namely 
\begin{equation*}
\frac{\partial L}{\partial p_1} \;=\; 
-\frac{T}{3p_0 + p_1}, 
\end{equation*}
which is always negative (since $p_0, p_1, T > 0$) and hence closest to 0 when $p_1 = 1$. Taking derivatives with respect to $p_0$ yields
\begin{equation*}
\frac{\partial L'}{\partial p_0} \;=\;
 -T\frac{3p_0 + p_1 + 3p_0 - 3}{p_0(3p_0 + p_1)},
\end{equation*}
which is $0$ when $p_0 = \frac{1}{3}$. Substituting these values back into the global predictions, we obtain that  $\mathbb{P}_{1 | I_1}(1 | I_1) = \frac{1}{2}$ for class $1$ images $I_1$ and $\mathbb{P}_{1 | I_0}(1 | I_0) = \frac{1}{3}$ for class $0$ images $I_0$. Hence, with a rounding threshold for class $1$ of $.5 + \epsilon$, at convergence, our method would predict that all images belong to class $0$.  However, even though the resulting accuracy would only be $50\%$, the feature heatmap (constructed as described in Appendix ~\ref{extracting_visualizations}) would clearly indicate that there is a feature representative of class $1$ in the upper left corner of the image, since $\mathbb{Q}_{1 | P^{(j)}}(1 | P^{(j)}) = 1$ for all patches $P^{(j)}$ containing only $1$-valued pixels.  Thus, the model is forced to identify every patch indicative of class $1$ during training as opposed to other models such as CAM and Grad-CAM that can rely on a single region for correct classification.   

This example explains mathematically why there is a trade-off between generalization error and granularity of feature detection and why PatchNet identifies as many features indicative of class $1$ and class $0$ in an image as possible. 

We now generalize our analysis form the example.  In particular, we analyze how the value of $\mathbb{Q}_{P}$ for each patch $P$ is related to the fraction of features indicative of class $1$ present in class $1$ images, the fraction of features indicative of class $0$ present in class $0$ images, and the fraction of features shared among both classes.  Intuitively, the model will assign low $\mathbb{Q}_{P}$ outputs for each patch $P$ with a feature indicative solely of class $0$, high $\mathbb{Q}_{P}$ outputs for each patch $P$ with a feature indicative solely of class $1$, and roughly $\mathbb{Q}_{P} = .5$ for patches with features shared between class $0$ and $1$.  In this section, we perform a case analysis on simplified images that contain only features that are indicative of class $1$ or features that are shared between class $0$ and class $1$. We mathematically determine the exact values of $\mathbb{Q}_{P}$ for each patch after convergence.

Due to the nature of our loss function, we can mathematically analyze the behavior of the model to understand the relationship between the visual  map and presence of class 1 features in the dataset under some simplifying assumptions.  Namely, suppose that the images in the dataset contain only features, $f_c$, common to both classes of the image or features, $f_1$, indicative of class $1$.  Now suppose that on average there are $a$ patches in class $1$ that are indicative of $f_c$ and $b$ patches in class $1$ that are indicative of $f_1$, and that all $a + b$ patches of class $0$ are indicative of $f_c$ with $a > b$.  Let us further assume that each patch contains only one of $f_c$ or $f_1$, and so $\mathbb{Q}_{1 | P^{(j)}}(1 | P^{(j)}) = p_0$ for all $j$ such that patch $P^{(j)}$ contains $f_c$ and $\mathbb{Q}_{1 | P^{(j)}}(1 | P^({j})) = p_1$ for all $j$ such that $P^{(j)}$ contains $f_1$.  Then, at convergence the model minimizes the loss

\begin{equation*}
\begin{split}
L &= -\log \left(1 - \frac{(a + b)p_0}{(a + b)} \right) -\log \left( \frac{ap_0 + bp_1}{a + b} \right) \\
&= -\log(1-p_0) - \log \left( \frac{ap_0 + bp_1}{a + b} \right) \\
\end{split}
\end{equation*}

In order to minimize the loss, we set the derivatives with respect to $p_0$ and $p_1$ equal to $0$. The derivative with respect to $p_1$ is:
\begin{equation*}
\frac{\partial L}{\partial p_1} =  - \frac{a  + b}{ap_0 + bp_1} \frac{b}{a + b}  = - \frac{b}{ap_0 + bp_1} \\
\end{equation*}
This derivative is always less than $0$ as  $a, b, p_1, p_0 \geq 0$.  Hence, the model will simply try to maximize the value of $p_1$ to bring the derivative closer towards $0$.  Thus, at convergence, $p_1 = 1$. Now examining the derivative with respect to $p_0$:
\begin{equation*}
\begin{split}
\frac{\partial L}{\partial p_0} &= 0 \\
\implies -\frac{1}{p_0 - 1} - \frac{a  + b}{ap_0 + bp_1} \frac{a}{a + b}&= 0\\
\implies ap_0 + bp_1 + ap_0 - a = 0 \\
\implies ap_0 + b + ap_0 - a = 0 \\ 
\implies p_0 = \frac{a - b}{2a} \\
\end{split}
\end{equation*}
Hence the largest possible value for $p_0$ is $\frac{1}{2}$, which occurs when $b = 0$.  Note that we performed this analysis for $a > b$.  If $a \leq b$, as $p_0$ is constrained to be in the range $\left[0, 1 \right]$, the value of $p_0$ will simply be $0$ at convergence.

As a concrete example of this convergence behavior, we examine the behavior of the model on a sample binary data set where class $0$ consists of all black $128 \times 128$ images (each pixel has value $0$), and class $1$ consists of $128 \times 128$ images with a white square of size $64 \times 64$ in the upper left hand corner (each pixel has value $1$) (a larger version of the sample image in Section 3.3). We train our model using $17 \times 17$ patches and a stride of $17$ in either direction.  In this case, there are some patches that contain both $f_c$ and $f_1$. Yet as there are only a few such patches, the actual values of $p_0$ and $p_1$ should only be slightly noisy.  We use zero padding and a stride size of $1$ to visualize the outputs of $\mathbb{Q}$ on each of the $128 \cdot 128 = 16384$ patches centered at each pixel value in the original image.  Since the $f_1$ features are contained in a $64 \times 64$ square in the upper left corner of the image, we claim that there are approximately $b = 64 \cdot 64 = 4096$ patches that contain $f_1$. Now, by our calculations, we expect that $p_0 = \frac{16384 - 2 \cdot 4096}{2 \cdot (16384 - 4096)} = \frac{1}{3}$ and that $p_1 = 1$.  That is, we expect our model to output a value of $\frac{1}{3}$ for patches containing only $f_c$ and $1$ for patches containing only $f_1$.  Indeed, after our model had converged, the model output a value of $0.333$ for patches consisting of only $f_c$ and output a value of $0.982$ for a patch consisting of only $f_1$.

\section{Extracting Visualizations}
\label{extracting_visualizations}
We now present how our method can easily provide visualizations of the learned features.  We refer to the gray-scale visualizations we generate as ``heatmap" visualizations, since brighter pixels in the visualizations indicate that the model found a feature relevant to class $1$, while darker pixels indicate a feature relevant to class $0$.  We refer to our output visualization as a global heatmap.

The \emph{global heatmap visualization} for an image is constructed by first computing $\mathbb{Q}_{1 | P^{(i,j)}}(1 | P^{(i,j)})$ for all patches $P^{(i,j)}$, where $P^{(i, j)}$  is the patch centered at location ${(i, j)}$ in the first two dimensions of the original image (zero-padding is used for border locations), and then constructing an $m \times n$ image where each pixel at location $(i', j')$ of the image is the value of $\mathbb{Q}_{1 | P^{(i',j')}}(1 | P^{(i',j')})$.  

As an aside, a different approach for constructing global heatmap visualizations is to average the predicted heatmap pixel values across all patches containing the given pixel.  We found that this approach provided empirically inferior results, in particular in terms of heatmap smoothness.

\section{Training Practices and Hyperparameter Selection}
\label{training_practices}
We adjusted the CAM and Grad-CAM models for our applications as follows. Since our images are smaller than 224 x 224, the size of images from ImageNet, we built a CAM model with a VGG-19 backbone with only 3 pooling layers. 
We built a Grad-CAM model using only a VGG-11 architecture instead of VGG-19, since the linear layers significantly overfit on some of our smaller datasets. CAM did not have this issue since it only makes use of the convolutional layers and not the linear layers. 
In order to train the VGG models for Grad-CAM, we re-sized all images to 224 x 224 pixels and all models were trained from scratch.

We use the following conventions for all visualizations: white regions indicate class $1$, whereas black regions indicate class $0$.  To obtain visualizations that are consistent with PatchNet, we provided CAM and Grad-CAM class $1$ labels so that white regions are indicative of features of class~$1$.

All models were trained using the Adam optimizer \citep{Adam} with a learning rate of $10^{-4}$, and we used Kaiming normal initialization~\citep{KaimingInit} for all convolutional layers.  The PatchNet subnetwork $\mathcal{S}$ was trained using a batch size of 4 (due to hardware memory limitations), while the CAM and Grad-CAM models were trained using a batch size of 64. To reduce overfitting, for all models we used a patience strategy~\citep{DeepLearning}, where we declared convergence when the model had not seen any improvement in validation loss for $1000$ epochs or when training accuracy reached 100\%.

\section{Describable Textures Dataset (DTD)}
\label{DTD_experiments}
DTD~\citep{Textures} is a collection of real-world texture images annotated with ``human-centric'' attributes.  The full dataset consists of 47 classes of textures with $120$ images per class. To be able to evaluate how well the different models identify the relevant features from the heatmap, we chose two classes where the features are obvious, namely \emph{cracks} versus \emph{perforations}. We assigned the images with cracks to class 0 and the images with perforations to class 1. We resized the images to 224 x 224 pixels in order to be able to train all models. We used $80$ images from each class for training and the remaining $40$ images from each class for validation, and we applied random flips to augment the training data.  

\begin{table}
\footnotesize
\centering
\begin{tabular}{|>{\centering}m{1.5cm}|>{\centering}m{1.5cm}|>{\centering}m{1.5cm}|>{\centering}m{1.5cm}|>{\centering\arraybackslash}m{1.5cm}|}
\hline
Model & Training Loss & Training Accuracy & Validation Loss & Validation Accuracy \\
\hline
PatchNet-31 & 0.091 & 99.4\% & 0.303 & 90.0\% \\
\hline
CAM & 0.277 & 91.8\% & 0.428 & 78.8\% \\
\hline
VGG-11 & 0.341 & 85.6\% & 0.516 & 75.0\%\\
\hline
\end{tabular}
\caption{Training and validation losses and accuracies for PatchNet-31, CAM, and VGG-11 used to construct Grad-CAM.  PatchNet-31 is able to achieve a much better validation loss by exploiting the repeated patterns in the images.}
\label{Cracked_Perforated_Acc_Table}
\end{table}

\begin{table}
\footnotesize
  \centering
\begin{tabular}{|>{\centering}m{1.9cm}|>{\centering}m{1cm}|>{\centering}m{1cm}|>{\centering}m{1cm}|>{\centering\arraybackslash}m{1cm}|}
\hline
Model & \multicolumn{2}{c|}{\parbox{2cm}{\centering Cracks}} & \multicolumn{2}{c|}{\parbox{2cm}{\centering Perforations}}  \\
\hline
Original &
\adjustimage{height=1cm,valign=m}{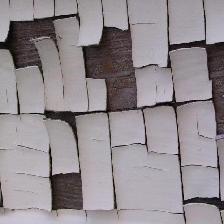} & 
\adjustimage{height=1cm,valign=m}{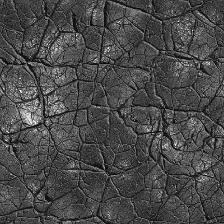} &
\adjustimage{height=1cm,valign=m}{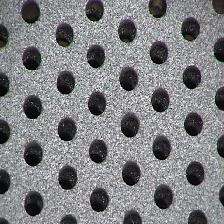} & 
\adjustimage{height=1cm,valign=m}{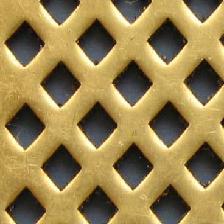} \\
\hline
PatchNet-31 &
\adjustimage{height=1cm,valign=m}{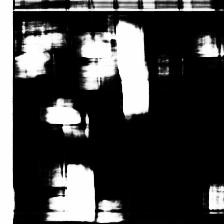} & 
\adjustimage{height=1cm,valign=m}{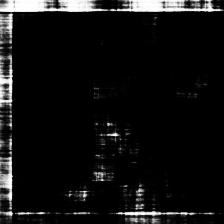} &
\adjustimage{height=1cm,valign=m}{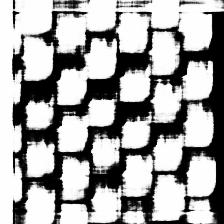} & 
\adjustimage{height=1cm,valign=m}{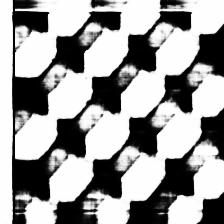} \\
\hline
CAM &
\adjustimage{height=1cm,valign=m}{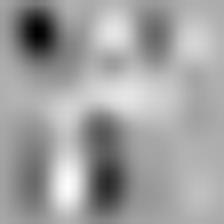} & 
\adjustimage{height=1cm,valign=m}{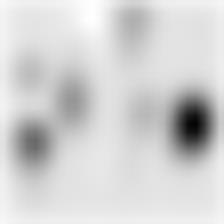} &
\adjustimage{height=1cm,valign=m}{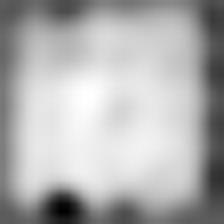} & 
\adjustimage{height=1cm,valign=m}{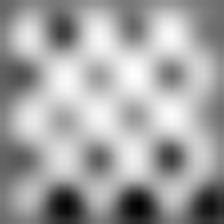} \\
\hline
Grad-CAM &
\adjustimage{height=1cm,valign=m}{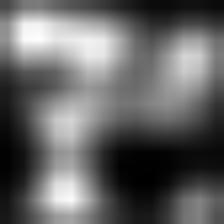} & 
\adjustimage{height=1cm,valign=m}{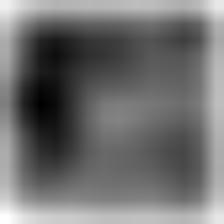} &
\adjustimage{height=1cm,valign=m}{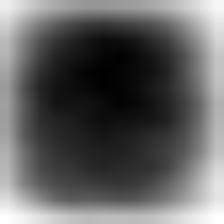} & 
\adjustimage{height=1cm,valign=m}{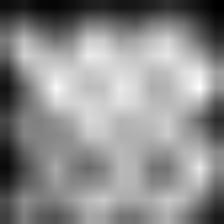} \\
\hline
\end{tabular}
\caption{Visualizations from PatchNet-31, CAM, and Grad-CAM.  PatchNet-31 provides much sharper visualizations by avoiding the need for an up-sampling procedure.}
\label{Cracked_Perforated_Masks}
\end{table}

For this application, PatchNet was trained with a patch size of 31 (PatchNet-31).
The resulting accuracies are shown in Table~\ref{Cracked_Perforated_Acc_Table}, indicating that PatchNet converged to a significantly better solution than CAM or Grad-CAM. As apparent in this application, using PatchNet as compared to CAM or Grad-CAM is particularly advantageous in small datasets where the features that are indicative for each class appear multiple times across the image. In such applications, PatchNet is able to converge to a better solution than other deep architectures, since its subnet can train on multiple patches per image, which due to the multiple appearances of the indicative features can be interpreted as a form of data augmentation.This significant difference in classification accuracies is less pronounced in datasets where the indicative features for each class do not tile the entire image. 

In Table~\ref{Cracked_Perforated_Masks}, we show examples of the resulting global heatmaps from PatchNet-31 together with visualizations from CAM and Grad-CAM. These examples show that all three models learned to identify the perforated regions in the images (represented as white marks in the visualizations of the three models). Interestingly, note that all models picked up the seemingly perforated regions in the left example of cracked images and highlighted them as perforations.   

As seen in these examples, PatchNet-31 provides sharper heatmaps than CAM or Grad-CAM on this dataset.  The limitations of CAM and Grad-CAM discussed in Section~\ref{intro_and_rel_work} are apparent in these visualizations:  Due to upscaling the last layer of filters, the CAM and Grad-CAM visualizations show a significant blurring effect. In addition, it is interesting to compare the visualizations provided by PatchNet-31 and Grad-Cam on the left example of perforated images. While PatchNet-31 highlighted every perforation as a class 1 feature, Grad-CAM only highlighted the outer-most perforations, thereby indicating that it made use of global context for the classification. 

\section{Classification between Normal and Fibrocystic Cell Nuclei}
\label{bj_mcf10a_experiments}

In Appendix \ref{DTD_experiments}, we chose a patch size of 31, since this allowed PatchNet to achieve a low validation loss while still providing sharp heatmap visualizations. In this application, we turn to experiments of medical relevance. We analyze the tradeoff between generalization error and sharpness of heatmap visualizations as patch size increases for the classification of cell nuclei based on DNA-stained images from \citep{Adit}. DNA condensation changes is one of the main features used by pathologists for cancer diagnosis~\citep{NanoscaleAlteration}. We here compare normal (BJ) versus fibrocystic (MCF10A) cell lines. Fibrocystic changes in a tissue represent benign or pre-malignant cancer states~\citep{IsotropicMorphometry}. It is hence important to determine indicative features of such cells to aid pathologists in early cancer diagnosis.

The dataset analyzed here contains 128 x 128 images of 1267 BJ cell nuclei and 1282 MCF10A cell nuclei. We used 190 images from each class for validation and all the remaining 2169 images for training. The BJ images were assigned to class $0$ and the MCF10A images to class $1$. To analyze the trade-off between generalization error and feature visualization, we trained three PatchNet subnets with patch sizes of 11, 17, and 31, respectively. 

\begin{table*}[!t]
\footnotesize
  \centering
\begin{tabular}{|>{\centering}p{2cm}|>{\centering}p{1.5cm}|>{\centering}p{1.5cm}|>{\centering}p{1.5cm}|>{\centering}p{1.5cm}|>{\centering\arraybackslash}p{4cm}|}
\hline
Model & Training Loss & Training Accuracy & Validation Loss & Validation Accuracy & Training Time Per Epoch (Seconds) \\
\hline
PatchNet-11 & 0.439 & 92.7\% & 0.553 & 80.8\% & 212.81 \\
\hline
PatchNet-17 & 0.381 & 92.6\% & 0.504 & 81.8\% & 131.63\\
\hline
PatchNet-31 & 0.226 & 95.4\% & 0.416 & 83.2\% & 79.22\\
\hline
CAM & 0.325 & 85.9\% & 0.456 & 82.1\% & 13.66\\
\hline
VGG-11 & 0.110 & 95.8\% & 0.242 & 90.5\% & 41.94\\
\hline
\end{tabular}
\caption{Tradeoff between patch size, generalization error (validation loss), and training time per epoch for PatchNet models as well as generalization error and training time per epoch for the CAM and VGG-11 models.  As patch size increases, generalization error decreases, and VGG-11 achieves lowest generalization error as expected for this larger dataset.}
\label{BJ_MCF10A_Acc_Table}
\end{table*}

Table~\ref{BJ_MCF10A_Acc_Table} shows the generalization error for the three PatchNet models, as well as for the CAM and VGG-11 model used for Grad-CAM. As expected, as patch size increases, the validation loss for PatchNet decreases, validation accuracy increases, and the time taken per epoch decreases. In this application, Grad-CAM achieved the lowest validation loss, which is expected since it uses the full VGG-11 architecture. Interestingly, PatchNet with a patch size of 31 achieved a lower validation loss than CAM on this dataset, although CAM can make use of the global image context. We believe that the reason for this low loss is, as described in the previous section, due to the fact that the sub-network in PatchNet could effectively use multiple patches per image to train on, while CAM could only train on entire images and hence PatchNet could converge to a better solution.

\begin{table}[!b]
\footnotesize
  \centering
\begin{tabular}{|>{\centering}m{1.2cm}|>{\centering\arraybackslash}m{1cm}|>{\centering\arraybackslash}m{1cm}|>{\centering}p{1cm}|>{\centering\arraybackslash}m{1cm}|}
\hline
Model & \multicolumn{2}{c|}{\parbox{2cm}{\centering \vspace{.7mm} Normal (BJ) Nuclei \vspace{.7mm} }}& \multicolumn{2}{c|}{\parbox{2cm}{\centering \vspace{.7mm} Abnormal \\ (MCF10A) Nuclei \vspace{.7mm}}} \\
\hline
Original  &
\adjustimage{height=.8cm,valign=m}{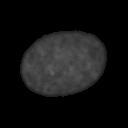} &
\adjustimage{height=.8cm,valign=m}{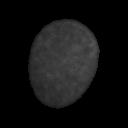} &
\adjustimage{height=.8cm,valign=m}{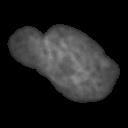} & 
\adjustimage{height=.8cm,valign=m}{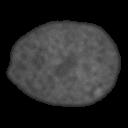} \\
\hline 
PatchNet-11 &  
\adjustimage{height=.8cm,valign=m}{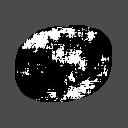} &
\adjustimage{height=.8cm,valign=m}{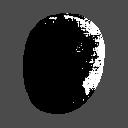} &
\adjustimage{height=.8cm,valign=m}{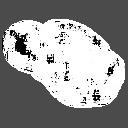} & 
\adjustimage{height=.8cm,valign=m}{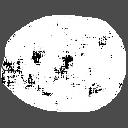} \\
\hline
PatchNet-17 & 
\adjustimage{height=.8cm,valign=m}{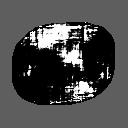} &
\adjustimage{height=.8cm,valign=m}{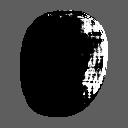} &
\adjustimage{height=.8cm,valign=m}{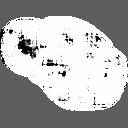}  & 
\adjustimage{height=.8cm,valign=m}{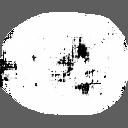}\\
\hline
PatchNet-31 & 
\adjustimage{height=.8cm,valign=m}{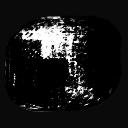} &
 \adjustimage{height=.8cm,valign=m}{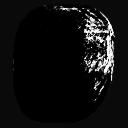} &
 \adjustimage{height=.8cm,valign=m}{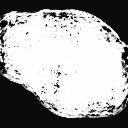} &
\adjustimage{height=.8cm,valign=m}{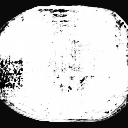}\\
\hline 
CAM  & 
\adjustimage{height=.8cm,valign=m}{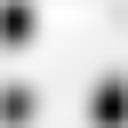} &
\adjustimage{height=.8cm,valign=m}{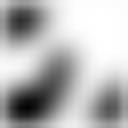} &
\adjustimage{height=.8cm,valign=m}{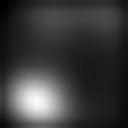} & 
\adjustimage{height=.8cm,valign=m}{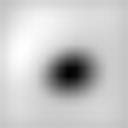} \\
\hline
GradCAM & 
\adjustimage{height=.8cm,valign=m}{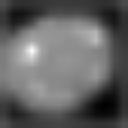} &
 \adjustimage{height=.8cm,valign=m}{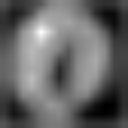} &
 \adjustimage{height=.8cm,valign=m}{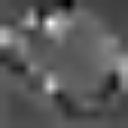} &
\adjustimage{height=.8cm,valign=m}{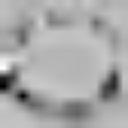} \\
\hline
\end{tabular}
\caption{Visualizations from PatchNet, CAM, and Grad-CAM for normal (BJ) and fibrocystic (MCF10A) cell nuclei.  While for all patch sizes, PatchNet provides relevant features, 
the CAM and Grad-CAM heatmaps mostly identify background as indicative features. 
}
\label{BJ_MCF10A_Vis_Table}
\end{table}

In Table~\ref{BJ_MCF10A_Vis_Table}, we provide the global heatmaps obtained from the three PatchNet models as well as the visualizations obtained from CAM and Grad-CAM for two randomly sampled nuclei of each class from the validation set.  It is apparent from these examples that the PatchNet heatmaps are significantly sharper for all patch sizes than the visualizations for the CAM and Grad-CAM models. The CAM and Grad-Cam visualizations again show a blurring effect. But most importantly, although Grad-CAM was able to achieve the lowest validation error, the visualizations for both, CAM and Grad-CAM are ineffective at highlighting regions of interest. 
For fibrocystic nuclei (class $1$), CAM  and Grad-CAM simply identify most of the image (including the background) as indicative of the class, while for normal cell nuclei (class $0$) the two models highlight the background as most indicative feature (since black regions in the visualization indicate features of class 0 and are found only in the background in most CAM and Grad-CAM visualizations for normal cell nuclei). 

Analyzing the global heatmaps from PatchNet for the different patch sizes together with the classification accuracies in Table~\ref{BJ_MCF10A_Acc_Table} shows the tradeoff between quality of global heatmaps and generalization error.  For small patch sizes, although there is a larger generalization error, the models provide sharper and more-localized feature heatmaps for each class as compared to large patch size, where the generalization error is lower, but the identified features show less granularity. Interestingly, PatchNet seems to pick up brighter and more granular regions in the original images as features for MCF10A cell nuclei. Fibrocystic cell nuclei are known to have more heterochromatin regions (corresponding to bright regions in the original images) and a clumpier texture (i.e., chromatin organization)~\citep{NanoscaleAlteration, IsotropicMorphometry}, which is in concordance with the features detected by PatchNet.

The fibrocystic state as well as cancer onset are strongly associated with an increase in DNA condensation~\citep{IsotropicMorphometry}. Heterochromatin regions (i.e., highly condensed DNA regions) recruit the histone \emph{CENP-A} and hence a fluorescent mark for CENP-A can be used as an indicator for DNA condensation. We note that DNA condensation occurs also in normal cells, but to a less extent than in MCF10A cells. We analyzed 354 normal (BJ) cell nuclei that were stained with CENP-A and overlayed the resulting masks with the features indicative of fibrocystic cell nuclei as determined by PatchNet-11, CAM and Grad-CAM. This allowed us to test whether the features identified by these models are related to DNA condensation and hence provide a functional annotation of the features.

\begin{table*}[t]
\footnotesize
  \centering
\begin{tabular}{|>{\centering}m{2cm}|>{\centering}m{1cm}|>{\centering}m{1cm}|>{\centering}m{1cm}|>{\centering}m{1.5cm}|>{\centering\arraybackslash}m{1.5cm}|>{\centering\arraybackslash}m{1.5cm}|}
\hline
Model & \multicolumn{3}{c|}{Examples} & Average Exact Match & Average Recall & Average AUROC \\
\hline
Original & 
\adjustimage{height=.8cm,valign=m}{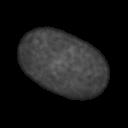} & 
\adjustimage{height=.8cm,valign=m}{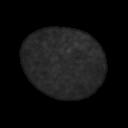} &
\adjustimage{height=.8cm,valign=m}{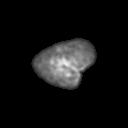} &
- & - & - \\
\hline
CENP-A Feature Masks & 
\adjustimage{height=.8cm,valign=m}{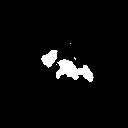} & 
\adjustimage{height=.8cm,valign=m}{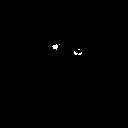} &
\adjustimage{height=.8cm,valign=m}{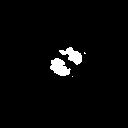}  &
- & - & - \\
\hline
PatchNet-11 & 
\adjustimage{height=.8cm,valign=m}{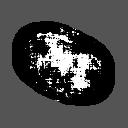} & 
\adjustimage{height=.8cm,valign=m}{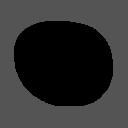} &
\adjustimage{height=.8cm,valign=m}{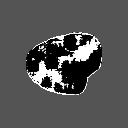} &
52.4\% & 50.4\% & .515 \\
\hline
CAM & 
\adjustimage{height=.8cm,valign=m}{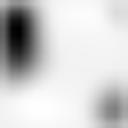} & 
\adjustimage{height=.8cm,valign=m}{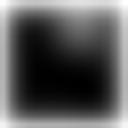} &
\adjustimage{height=.8cm,valign=m}{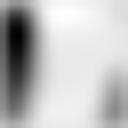} &
28.9\% & 89.1\% & .452 \\
\hline
Grad-CAM & 
\adjustimage{height=.8cm,valign=m}{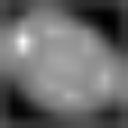} & 
\adjustimage{height=.8cm,valign=m}{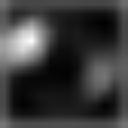} &
\adjustimage{height=.8cm,valign=m}{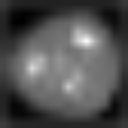} &
32.2\% & 78.2\% & .473 \\
\hline
\end{tabular}
\caption{PatchNet, CAM, and Grad-CAM visualizations are compared with CENP-A annotations of DNA condensation, a hallmark of cancer onset, on normal cell nuclei after only training the models to distinguish between normal (BJ) and fibrocystic (MCF10A) nuclei.  The features identified by PatchNet are correlated with the CENP-A annotations, while CAM and Grad-CAM return biologically irrelevant features and classify most of the nucleus as fibrocystic.}
\label{BJ_MCF10A_Masks}
\end{table*}

After generating the global heatmaps using our models on these 354 BJ nuclei, we computed the following statistics for each image:
\begin{itemize}
\item \emph{Average Exact Match:} average over all images of the number of pixel values that matched exactly between the CENP-A mask and the visualization heatmaps rounded at a threshold~of~0.5.  
\item \emph{Average Recall:} average of the recall per image across all images in the dataset, i.e., the number of CENP-A pixels that were identified by the model's heatmap (rounded at a threshold of 0.5) as being indicative of MCF10A divided by the total number of CENP-A pixels present in the mask.
\item \emph{Average AUROC:} average over all images of the area under the receiver operating characteristic curve.
\end{itemize}
The rationale for choosing these statistics is as follows: Since DNA condensation is associated with cancer onset, we used recall to measure the proportion of DNA condensation indicators that were picked up by the different models.  However, a high recall can be achieved by highlighting the whole nucleus as being indicative of MCF10A. Hence, we used exact match to determine the percentage of the nucleus that was tagged correctly as being normal.  To compare how classification accuracies vary with different rounding thresholds, we used the AUROC metric.  We chose not to analyze precision, since a high precision can only be achieved if CENP-A markers were the only significant features for cancer classification, which is not the case.  

Table~\ref{BJ_MCF10A_Masks} provides the average exact match, average recall, and sample heatmaps for the PatchNet-11, CAM and Grad-CAM models. Since all statistics are higher than 50\% for PatchNet, this indicates that the identified features are indeed related to DNA condensation. Note that CAM and Grad-CAM both achieve high average recall. However, as shown by the low average exact match and the visualizations in Table~\ref{BJ_MCF10A_Vis_Table}, this is due to the fact that these models often identify the whole nucleus as being indicative of MCF10A.

\section{Classification between Human and Mice Cells}
\label{bj_nih_nuclei}
Here, we present an additional experiment on cell nucleus classification using PatchNet, CAM, and Grad-CAM models.  Namely, we provide a baseline experiment to distinguish between human (BJ) and mice (NIH/3T3) cells.  The dataset contains 128 x 128 zero-padded images of 557 human and 557 mouse cell nuclei \citep{Adit}. We used 475 images from each class for training and the remaining 82 images from each class for validation.  The human cell nuclei were assigned to class $0$ and the mouse cell nuclei to class $1$.  This can be considered a simple classification task, since mouse cell nuclei are easily distinguishable from human cell nuclei due to their bright heterochromatin spots (i.e., regions of condensed DNA). To further establish that this experiment is a simple classification task, we use a logistic regression that achieves 100.0\% training and 96.3\% validation accuracy when using the following features from \citep{Mahotas}:  
\begin{itemize}
\item Area of the nucleus
\vspace{-0.1cm}
\item Eccentricity of the nucleus
\vspace{-0.1cm}
\item Roundness of the nucleus
\vspace{-0.1cm}
\item Linear Binary Patterns using 8 points and a radius of 4 \citep{LBP}
\vspace{-0.1cm}
\item Parameter-Free Threshold Adjacency Statistics \citep{PFTAS}
\end{itemize}

To analyze the trade-off between generalization error and feature visualization, we trained three PatchNet subnets with patch sizes of 11, 17, and 31, respectively.  Since the nucleus images were zero-padded and in order to ensure that the subnets were trained on few all-zero patches, we only used the patches from the central 64 x 64 image region.  Table~\ref{BJ_NIH_Acc_Table} shows the generalization error for the three PatchNet models, as well as for the CAM and VGG-11 model used for Grad-CAM. As expected, as patch size increases, validation loss for PatchNet and time taken per epoch decreases.  We note that possibly counterintuitively, validation accuracy for the PatchNet models decreases as patch size increases. We verified that this is due to fluctuations in the sigmoid values around the 0.5 classification threshold for a few difficult-to-classify images. In this application, CAM achieves the lowest validation loss, which can be explained by the fact that contrary to PatchNet this model can take advantage of the global image context for classification and contrary to Grad-CAM it does not overfit the data. The VGG-11 model used for Grad-CAM overfits the data primarily due to the final set of linear layers and the small amount of data used for training.

\begin{table*}[!t]
\footnotesize
  \centering
\begin{tabular}{|>{\centering}p{2cm}|>{\centering}p{1.5cm}|>{\centering}p{1.5cm}|>{\centering}p{1.5cm}|>{\centering}p{1.5cm}|>{\centering\arraybackslash}p{4cm}|}
\hline
Model &  Training Loss & Training Accuracy & Validation Loss & Validation Accuracy & Training Time Per Epoch (Seconds)\\
\hline
Patchnet-11 & 0.127 & 98.6\% & 0.154 & 98.2\% & 24.41 \\
\hline
Patchnet-17 & 0.050 & 99.6\% & 0.104 & 97.6\% & 16.27 \\
\hline
Patchnet-31 & 0.004 & 99.9\% & 0.074 & 97.0\% & 10.03 \\
\hline
CAM & 0.014 & 99.4\% & 0.027 & 98.8\% & 6.04 \\
\hline
VGG-11 & 0.042 & 98.1\% & 0.138 &  97.0\% & 4.63 \\
\hline
\end{tabular}
\caption{Tradeoff between patch size, generalization error (validation loss), and training time per epoch for PatchNet models as well as generalization error and training time per epoch for the CAM and VGG-11 models.  As patch size increases, generalization error decreases. CAM achieves the lowest generalization error as it has access to global context and doesn't overfit as much as VGG.}
\label{BJ_NIH_Acc_Table}
\end{table*}

In Table~\ref{BJ_NIH_Vis_Table}, we provide the global heatmaps obtained from the three PatchNet models as well as the visualizations obtained from CAM and Grad-CAM for two randomly sampled nuclei of each class from the validation set.  Again, PatchNet heatmaps are sharper for all patch sizes than the visualizations for the CAM and Grad-CAM models. The CAM and Grad-CAM visualizations again show a blurring effect. But again, although Grad-CAM was able to achieve the lowest validation error, the visualizations for both, CAM and Grad-CAM are ineffective at highlighting regions of interest.
For mice cell nuclei (class $1$), CAM seems to highlight most of the image as being indicative of the class while Grad-CAM highlights just the background as indicative of the class.  As some mice cell nuclei are smaller than human cell nuclei, Grad-CAM is able to use the background to classify these nuclei correctly.  

Analyzing the results from Tables \ref{BJ_NIH_Acc_Table} and \ref{BJ_NIH_Vis_Table} shows the tradeoff between quality of global heatmaps and generalization error.  For small patch sizes, although there is a larger generalization error, the models provide sharper and more localized feature heatmaps for each class as compared to large patch size, where the generalization error is lower, but the identified features show less granularity (i.e., more of the nucleus is colored correctly since larger patches contain more context). However, we can see that our PatchNet models correctly learned high luminance contrast as a distinguishing feature of NIH/3T3 (mouse) cell nuclei since the edges of each nucleus with the 0 padded background are colored white.

\begin{table}[!t]
\footnotesize
  \centering
\begin{tabular}{|>{\centering}m{1.2cm}|>{\centering\arraybackslash}m{1cm}|>{\centering\arraybackslash}m{1cm}|>{\centering}m{1cm}|>{\centering\arraybackslash}m{1cm}|}
\hline
Model & \multicolumn{2}{c|}{\parbox{2cm}{\centering \vspace{.7mm} Normal (BJ) Nuclei \vspace{.7mm} }}& \multicolumn{2}{c|}{\parbox{2cm}{\centering \vspace{.7mm} Mouse \\ (NIH/3T3) Nuclei \vspace{.7mm}}} \\
\hline
Original  &
\adjustimage{height=.8cm,valign=m}{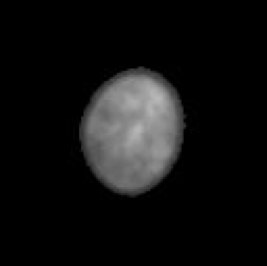} &
\adjustimage{height=.8cm,valign=m}{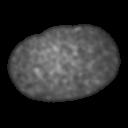} &
\adjustimage{height=.8cm,valign=m}{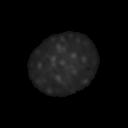} & 
\adjustimage{height=.8cm,valign=m}{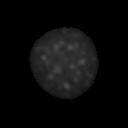} \\
\hline 
PatchNet-11 &  
\adjustimage{height=.8cm,valign=m}{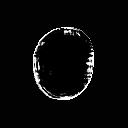} &
\adjustimage{height=.8cm,valign=m}{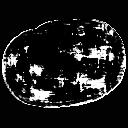} &
\adjustimage{height=.8cm,valign=m}{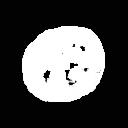} & 
\adjustimage{height=.8cm,valign=m}{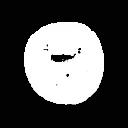} \\
\hline
PatchNet-17 & 
\adjustimage{height=.8cm,valign=m}{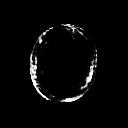} &
\adjustimage{height=.8cm,valign=m}{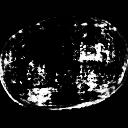} &
\adjustimage{height=.8cm,valign=m}{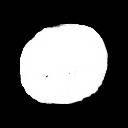}  & 
\adjustimage{height=.8cm,valign=m}{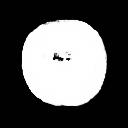}\\
\hline
PatchNet-31 & 
\adjustimage{height=.8cm,valign=m}{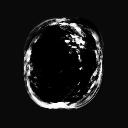} &
 \adjustimage{height=.8cm,valign=m}{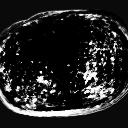} &
 \adjustimage{height=.8cm,valign=m}{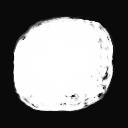} &
\adjustimage{height=.8cm,valign=m}{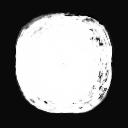}\\
\hline 
CAM  & 
\adjustimage{height=.8cm,valign=m}{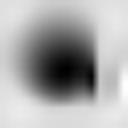} &
\adjustimage{height=.8cm,valign=m}{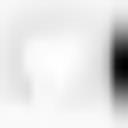} &
\adjustimage{height=.8cm,valign=m}{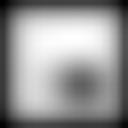} & 
\adjustimage{height=.8cm,valign=m}{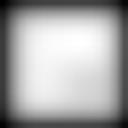} \\
\hline
Grad-CAM & 
\adjustimage{height=.8cm,valign=m}{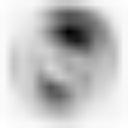} &
 \adjustimage{height=.8cm,valign=m}{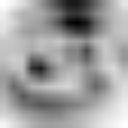} &
 \adjustimage{height=.8cm,valign=m}{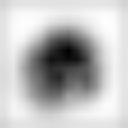} &
\adjustimage{height=.8cm,valign=m}{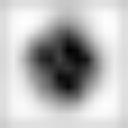} \\
\hline

\end{tabular}
\caption{Visualizations from PatchNet, CAM, and Grad-CAM for normal (BJ) and mouse (NIH/3T3) cell nuclei.  While for all patch sizes, PatchNet provides relevant features, 
the CAM and Grad-CAM heatmaps mostly identify background as indicative features. 
}
\label{BJ_NIH_Vis_Table}
\end{table}

\section{Hardware and Software Specifications}
We ran our models on PyTorch~\citep{PyTorch} 0.1.12 on Python 3.6 packaged under Anaconda 4.3.17 with Nvidia driver version 375.51, Cuda version 8.0.61 and CuDNN version 5.1.0. We used SciKit-Learn~\citep{SciKit-Learn}, Numpy~\citep{NumPy} and Mahotas~\citep{Mahotas} as tools while developing our models. We used Facebook's Visdom~\citep{Visdom} as a tool to visualize filters and display them for this paper.

For training, we used one server running Ubuntu 16.04.2 with an Intel i7-4930K at 3.40GHz with 56GB DDR3 RAM and two Nvidia Titan x (Pascal) with 12GB of GDDRX RAM.

\end{document}